\documentclass[10pt,twocolumn,letterpaper]{article}

\usepackage{cvpr}
\usepackage{times}
\usepackage{epsfig}
\usepackage{graphicx}
\usepackage{amsmath}
\usepackage{amssymb}
\usepackage{bbm}
\usepackage{dblfloatfix}    

\usepackage{algorithm}
\usepackage[]{algpseudocode}

\algnewcommand{\LeftComment}[1]{\Statex \textit{#1}}

\usepackage[pagebackref=true,breaklinks=true,letterpaper=true,colorlinks,bookmarks=false]{hyperref}

\cvprfinalcopy 

\def\cvprPaperID{****} 

\ifcvprfinal\fi
\begin{document}

\title{Semantic Segmentation Refinement by Monte Carlo Region Growing of High Confidence Detections}


\author{Philipe A. Dias\\
{\tt\small philipe.ambroziodias@marquette.edu}
\and
Henry Medeiros\\
{\tt\small henry.medeiros@marquette.edu}\\
EECE,   
        Marquette University\\
Milwaukee, WI, USA
}



\maketitle

\begin{abstract}
Despite recent improvements using fully convolutional networks, in general, the segmentation produced by most state-of-the-art semantic segmentation methods does not show satisfactory adherence to the object boundaries. We propose a method to refine the segmentation results generated by such deep learning models. Our method takes as input the confidence scores generated by a pixel-dense segmentation network and re-labels pixels with low confidence levels. The re-labeling approach employs a region growing mechanism that aggregates these pixels to neighboring areas with high confidence scores and similar appearance. In order to correct the labels of pixels that were incorrectly classified with high confidence level by the semantic segmentation algorithm, we generate multiple region growing steps through a Monte Carlo sampling of the seeds of the regions. Our method improves the accuracy of a state-of-the-art fully convolutional semantic segmentation approach on the publicly available COCO and PASCAL datasets, and it shows significantly better results on select sequence of the finely-annotated DAVIS dataset.
\end{abstract}

\section{Introduction}
Image understanding is a primary goal of computer vision, which is crucial to several applications ranging from activity recognition to object tracking, autonomous navigation, security and surveillance, among many others. One of the most important aspects of image understanding is the determination of the objects present in the image. Such determination can be carried out at different levels of granularity, which correspond to four well-known subproblems of image understanding: image classification, object detection, semantic segmentation, and instance segmentation. While the objective of image classification is to label the scene at the image-level and object detection attempts to localize the bounding boxes of known objects, segmentation tasks aim at pixel-wise classification of known objects or of different instances of these objects. These subproblems aim to achieve increasingly complex goals and have therefore been addressed with different levels of success.

In recent years, the combination of Deep Convolutional Neural Networks (DCNN) and increasingly larger publicly available datasets has led to substantial improvements in the state-of-the-art of image classification tasks \cite{Krizhevsky2012,He}. Similar advances became then possible for object detection with the advent of region proposal networks (RPN) \cite{Ren2015}. For segmentation tasks, however, the performance of typical DCNNs architectures is limited by the typical downsampling (pooling) intrinsic to such networks. Downsampling is employed to learn hierarchical features (from low-level to high-level features), but it ultimately leads to imprecise, coarse segmentation in scenarios that require pixel-wise predictions. Different strategies with varying levels of complexity have been proposed to address this limitation, including \textit{atrous} convolutions \cite{Chen2015}, upsampling strategies \cite{Long2015,Ronneberger,Lin2016}, and post-processing based on superpixels \cite{Mostajabi} or conditional random fields (CRFs) \cite{Chen2015,Krahenbuhl2012}. Although these methods represent significant steps towards solving this problem, the segmentation they produce tend not to be finely aligned with the boundaries of the objects.

In this paper, we propose the \textit{Region Growing Refinement (RGR)} algorithm, a post-processing module that operates on the principle of appearance-based region growing to refine the predictions generated by a DCNN for semantic segmentation. Based on the classification scores available from the detector, our method first divides the image into three regions: high confidence background, high confidence object, and uncertainty region. The pixels within the uncertainty region, which are the ones that tend to be misclassified by DCNN-based methods, are then labeled by means of region growing. We apply Monte Carlo sampling to select initial seeds from the high confidence regions, which are then grown using a similarity criterion based on a distance metric computed in the 5-D space of spatial and color coordinates. Our model has the advantage of not requiring any dataset-specific parameter optimization, working in a fully unsupervised manner.

We employ the  Fully Convolutional Instance-aware Semantic Segmentation (FCIS) algorithm \cite{Li2016}, the winner of the MS COCO 2016 Semantic Segmentation Challenge \cite{Lin2016}, as a predictor module as well as the baseline for our performance evaluation. When evaluating our results on the MS COCO dataset, we observed that the available ground truth annotations contain non-negligible inaccuracies in terms of adherence to the actual boundaries of the objects. To assess the effects of this problem and achieve a more accurate evaluation of the benefits of our approach, we also assess its efficacy on the PASCAL VOC2012 \cite{pascal2012} validation set as well as on selected video sequences from the DAVIS (Densely Annotated VIdeo Segmentation) dataset \cite{Perazzi2016}. Compared to the first two datasets, annotations provided for the DAVIS dataset are more fine-grained, with tighter adherence to actual object real boundaries. As a result, in addition to relatively small increases in segmentation accuracy for the MS COCO dataset ($+1.5\%$ in $AP$) and the PASCAL validation sets ($+0.5\%$ in $mAP\%$), we report significantly better quantitative results  for the DAVIS sequences ($+3.2\%$ in $IoU\%$), which more realistically reflect the segmentation improvement observed through qualitative (visual) inspection.


\section{Related Work}
The relatively recent development of DCNNs trained on large publicly available datasets represented an inflection point in image understanding research. Compared to the period when models based on hand-engineered features (e.g. HOG \cite{Dalal}, SIFT \cite{Lowe2004}) were the norm, currently the state-of-the-art in object recognition tasks is being improved at a dramatically faster pace. Between 2011 and 2015, the top-$5$ classification error on the ImageNet dataset was reduced from $25.8\%$ to $3.57\%$ \cite{He}. 

However, traditional CNN-based methods that are effective for image classification and object detection present limitations for segmentation tasks. The combination of max-pooling and \textit{striding} operations constitute a standard strategy for learning hierarchical features, which are a determining factor in the success of deep learning models. They explore the transition invariance favored by image-level labeling tasks, i.e., the fact that only the presence of an object matters, not its location. Such premise, however, does not hold for pixel-dense classification tasks, which require precise object localization. As pointed out by Chen et al. in \cite{Chen2015}, such spatial insensitivity and image downsampling inherent to conventional DCNNs are two major hindering factors for their performance in image segmentation. A particularly noticeable effect is that these methods generate coarse segmentation masks, with limited boundary adherence.

One particularly successful strategy investigated for segmentation tasks is the concept of fully convolutional networks (FCNs) \cite{Long2015}. In FCNs, the traditionally fully connected layers at the end of conventional CNNs are replaced by convolutional layers, which upsample the feature maps to generate a pixel-dense prediction. Different upsampling or \textit{decoding} approaches have been proposed, including the use of deconvolution layers \cite{Noh2015,Long2015}, and encoder-decoder architectures with skip-layer connections \cite{Ronneberger,Kendall,Hariharan2015,Lin2016}. While these approaches aim at recovering the information lost due to downsampling, other methods focus on reducing the downsampling rate. The Deeplab architecture \cite{Chen2015} is one such method in which the concept of dilated (or \textit{atrous}) convolutions is successfully employed to reduce the downsampling rate from $1/32$ to $1/8$. Such strategy, however, performs convolutions on high-resolution feature maps, requiring substantially more computational resources than other approaches.


A variation of the semantic segmentation problem is instance segmentation, which requires detecting and segmenting individual object instances. Coarse segmentations are a significant challenge for this type of task, since neighboring instances of objects of the same class are frequently merged into a single segmentation. Dai et al. in \cite{Dai2016} introduce the concept of instance-sensitive FCNs, in which an FCN is designed to compute score maps that determine the likelihood that a pixel belongs to a relative position (e.g.right-upper corner) of an object instance. FCIS \cite{Li2016} is an extension of this approach, which achieved the state-of-the-art performance and won the COCO 2016 segmentation competition.

In addition to adjustments in CNN architectures, multiple works have investigated the use of low-level information strategies to construct models for object detection and semantic segmentation. Following the success of CNNs on image classification tasks, Girschick et al. in \cite{Girshick2014} introduced the model known as RCNN that performs object detection by evaluating multiple region proposals. In RCNN, the region proposals are generated using Selective Search \cite{Uijlings2013}, which consists on merging a set of small regions \cite{Felzenszwalb} based on different similarity measures, an approach closely related to the concept of superpixels.

Superpixel are perceptually meaningful clusters of pixels, which are grouped based on color similarity and other low-level properties. Stutz et al. \cite{Stutz} provide a review of existing superpixel approaches, with a comprehensive evaluation of $28$ state-of-the-art algorithms. One of the most widely-used method is the Simple Linear Iterative Clustering (SLIC) algorithm \cite{Achanta2012}, a \textit{k-means}-based method that groups pixels into clusters according to their proximity in both spatial and color spaces. Semantic segmentation models using superpixels traditionally employ them as a pre-processing step. The image is first divided into superpixels, followed by a prediction step in which each element is individually evaluated using engineered hierarchical features \cite{Gould2008} or DCNNs \cite{Couprie2013,Mostajabi}.

In addition to the generation of region proposals or pre-segmentation elements, local-appearance information has been also employed in post-processing steps to improve segmentations obtained with deep CNN models. One such post-processing approach is to model the problem as a CRF defined over neighboring pixels or small patches, which however might lead to excessive smoothing of boundaries due to the lack of long-range connections \cite{Krahenbuhl2012}. Kr{\"a}henb{\"uhl} et al. in \cite{Krahenbuhl2012} introduces an efficient algorithm for fully connected CRFs containing pairwise potentials that associate all pairs of pixels in a image. This algorithm is successfully exploited by the Deeplab model \cite{Chen2015} to produce fine-grained segmentations. However, this refinement model contains parameters that have to be optimized in a supervised manner when applied to different datasets. 


\section{Proposed Approach}
\begin{figure*}[t]
  \centering
  \includegraphics[width=.98\textwidth]{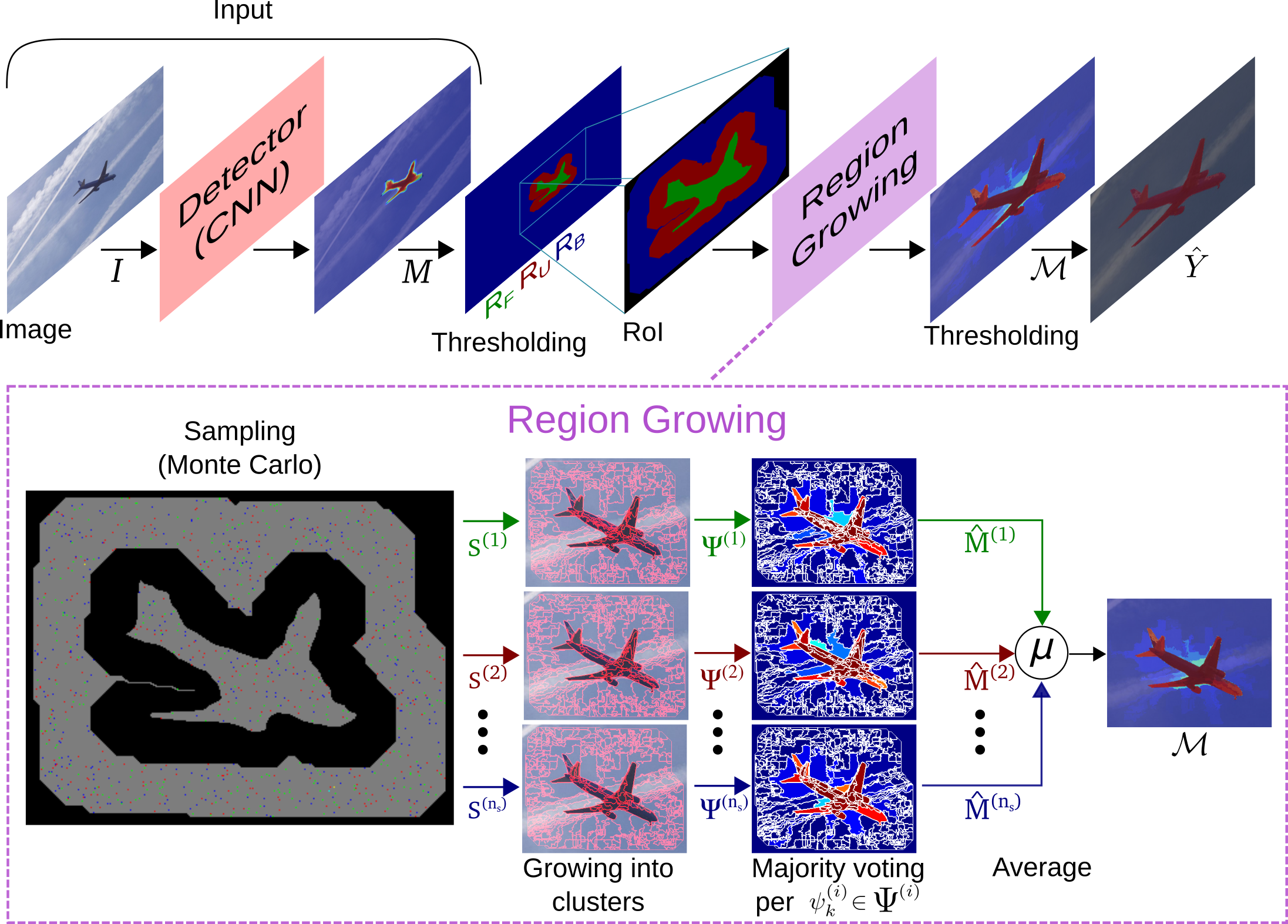}
    \caption{\textbf{Best viewed in color.} Diagram illustrating the sequence of tasks performed by the proposed RGR model for segmentation refinement. Each task and its corresponding output (shown above the arrows) are described in Algorithm \ref{alg:pseudocode}. }
    \label{fig:flow}
\end{figure*}

The method we propose for refinement of segmentation boundaries is a generic post-processing module that can be coupled to the output of any CNN or similar model for semantic segmentation. Our RGR algorithm consists of four main steps: 1) identification of low and high confidence classification regions; 2) Monte Carlo sampling of initial seeds; 3) region growing; and, 4) majority voting and final classification. The operations that comprise these steps are described in detail below. In our description, we make reference to Figure \ref{fig:flow} and Algorithm \ref{alg:pseudocode}, which lists the operations performed by our method for each proposed detection in an image. If multiple detections are present, the algorithm is executed once for each detection.

\textit{1) Step 1 - Thresholding the image into three regions:} Conventional models typically infer a final classification by pixel-wise thresholding the scores obtained for each class. In such case, the selection of an adequate threshold represents a trade-off between false positives and false negatives. Instead of relying on intermediate threshold values, our refinement module directly exploits the available classification scores to differentiate between three regions: a) high confidence foreground (or object) $R_{F}$; b) high confidence background $R_{B}$; and, c) uncertainty region $R_{U}$. As defined in Eq. \ref{eq:ru}, these regions are identified using two high confidence thresholds $\tau_F$ and $\tau_B$
\begin{alignat}{3}
R_{F}&=\left\{ p_{j}|M(p_{j})>\tau_{F}\right\}, \nonumber \\
R_{U}&=\left\{ p_{j}|\tau_B < {M}(p_j) < \tau_F\right\}, \label{eq:ru}  \\
R_{B}&=\left\{ p_{j}|{M}(p_j) < \tau_B\right\}, \nonumber
\end{alignat}
where $p_j$ is the $j$-th pixel in the input image $I$, and $M(p_j)$ is its corresponding detection score computed according to the detection confidence map $M$. High confidence regions correspond to areas where pixels present scores near the extremes of the likelihood range. For normalized scoremaps, values lower than a threshold $\tau_B \sim 0.0$ identify pixels in the high confidence background, while values larger than $\tau_F \sim 1.0$ indicate high confidence foreground elements. In order to recover possible false negatives, morphological thickening is performed on the background $R_B$ boundaries. The fourth leftmost image in Figure \ref{fig:flow} illustrates the division of the image into the three regions $R_F$, $R_U$, and $R_{B}$.

\textit{2) Step 2 - Monte Carlo sampling of initial seeds:} Inspired by the notion of pixel affinity employed by many algorithms for superpixel segmentation, our method applies a region growing approach to classify pixels within the uncertainty zone. More specifically, we build on the simple non-iterative clustering (SNIC) algorithm \cite{Achanta2017}.

The first difference between RGR and SNIC is the selection of initial seeds. SNIC selects initial seeds on a regular grid over the whole image. In contrast, our algorithm selects seeds by Monte Carlo sampling the high confidence regions. Such an adaptation guarantees that pixels composing the uncertainty region will be labeled according to high confidence information.

Our Monte Carlo approach consists of $n_s$ iterations. In each iteration $i$, a set $S^{(i)}$ of initial seeds is defined by uniformly sampling the high-confidence area $R_H=R_{B}\cup R_{F}$. Let $p_h$ represents pixels within the region $R_H$, where the index $h=1,\ldots,|R_H|$. Uniform sampling of seeds can thus be performed index-wise according to $h \sim U (1,|R_H|)$.

We determine the number of seeds to be sampled as a function of the sampling domain area (i.e. $|R_H|$) and the desired average spacing between samples. Such a spacing between seeds is necessary to ensure the availability of paths through which all initial centroids can propagate through the uncertainty region. By drawing multiple sets of seeds, we reduce the impact of possible high confidence false positives on the final segmentation.

\begin{algorithm}
  \caption{RGR refinement algorithm}\label{euclid}
  \label{alg:pseudocode}
  \begin{algorithmic}[1]
      \Require{Image $I$, confidence map $M$}
      \Ensure{Refined semantic segmentation $\hat{Y}$ of image $I$} 
      \State Threshold $M$ into three regions: background $R_B$, foreground $R_F$ and uncertain zone $R_U$  \label{ln:seg}
      \State Define a Region of Interest (RoI) according to Eq. \ref{eq:roi}
      \For {i = 1 to $n_s$}
      \State \parbox[t]{\dimexpr\linewidth-\algorithmicindent}{Form a set $S^{(i)}$ of initial seeds by uniformly sampling from $R_{B}\cup R_{F}$}
      \label{ln:crop}
      \State \parbox[t]{\dimexpr\linewidth-\algorithmicindent}{Generate a set of clusters $\Psi^{(i)}$ by performing region growing using the SNIC algorithm with $S^{(i)}$ as input} \label{ln:pad}    
	\State \parbox[t]{\dimexpr\linewidth-\algorithmicindent} {For each generated cluster $\psi^{(i)}_{k}\in\Psi^{(i)}$, compute confidence map $\hat{M}^{(i)}$ according to Eq. \ref{eq:majvoting}}
      \EndFor
      \State Compute the pixel-wise average $\mathcal{M}$ of $\hat{M}^{(i)}$, $i=1,\ldots,n$  \label{ln:svm}
      \State Generate $\hat{Y}$ by pixel-wise thresholding $\mathcal{M}$ 
  \end{algorithmic}
\end{algorithm}

\textit{3) Step 3 - Region Growing:} Initial centroids are grown according to an adaptation of the SNIC algorithm. The dashed block at the bottom of Figure \ref{fig:flow} illustrates the sequence of operations performed by RGR for region growing. Like its predecessor SLIC \cite{Achanta2012}, SNIC measures the similarity between a pixel and a centroid as their distance in a five-dimensional space of color and spatial coordinates. Let the spatial position of a pixel be represented by a vector $\mathbf{x} = [x\ y]^T$, while its color is expressed in the CIELAB color-space by $\mathbf{c} = [l\ a\ b]^T$. The distance between a pixel \textit{j} and a centroid \textit{k} is given by
\begin{equation}
\label{eq:distsnic}
d_{j,k}=\sqrt{\frac{\lVert x_{j}-x_{k}\rVert_{2}^{2}}{\theta_s}+\frac{\lVert c_{j}-c_{k}\rVert_{2}^{2}}{\theta_m}},
\end{equation}
where \textit{$\theta_s$} and \textit{$\theta_m$} as normalizing factors for spatial and color distances, respectively. 

Aiming at lower computational times, we restrict the region growing to a Region of Interest (RoI) around the uncertain zone $R_U$. Based on the spatial distance to $R_U$, the background $R_B$ can be split into two regions: far background $R_{fB}$ and near background $R_{nB}$. Since the far background is unlikely to influence the clustering of the uncertain region, $R_{fB}$ can be ignored during the region growing computation. Hence, we define the RoI for region growing as
\begin{equation}
\label{eq:roi}
	RoI = R_{nB} \cup R_{F} \cup R_{U}.
\end{equation}

The region growing implementation relies on a priority queue, which is constantly populated with nodes that correspond to unlabeled pixels 4 or 8-connected to a region being grown. This queue is sorted according to the similarity between the candidate pixel and the growing region, given by the metric defined in Eq. \ref{eq:distsnic}. While the queue is not empty, each iteration consists of: 1) popping the first element of the queue, which corresponds to the unlabeled pixel that is mostly similar to a neighboring centroid $k$; 2) annexing this pixel to the respective cluster $\psi^{(i)}_k$; 3) updating the region centroid; and, 4) populating the queue with neighbors of this pixel that are still unlabeled. 

Compared to the original SNIC algorithm, we add a constraint in order to reduce the incidence of false detections. A node is only pushed into the queue if its distance to the corresponding centroid is smaller than a certain value $d_{\text{Max}}$. In this way, we ensure that an unlabeled pixel within $R_U$ will only inherit information from high confidence pixels that are sufficiently similar to it. This creates the possibility of ``orphan'' elements, i.e., pixels for which no node is ever created. Such pixels are therefore classified as background. For each set of initial seeds $S_i$ the region growing process generates as output a cluster map $\Psi^{(i)}$ that associates each pixel to a respective cluster $\psi^{(i)}_{k}$.

\textit{4) Step 4 - Majority voting and final classification:} Following the region growing process, RGR conducts a majority voting procedure to ultimately classify each generated cluster into foreground or background. As expressed in Eq. \ref{eq:majvoting}, a pixel $p_j$ contributes a positive vote for foreground classification if its original prediction score $M(p_j)$ is larger than a threshold $\tau_0$. We compute the ratio of positive votes across all pixels $p_j\in\psi^{(i)}_{k}$, generating a refined likelihood map $\hat{M}^{(i)}$ for each set of clusters according to 
\begin{align}
\label{eq:majvoting}
Y&=\left\{p_{j}|M(p_{j})>\tau_{0}\right\}, \\
\hat{M}^{(i)}(p_{j})&=\frac{|Y|}{|\psi_{k}^{(i)}|}
\end{align}

The likelihood maps $\hat{M}_{(i)}$ obtained from the $n_s$ Monte Carlo samplings are averaged to generate a final pixel dense score map $\mathcal{M}=\frac{1}{n_s}\sum_{i=1}^{n_{s}}M^{(i)}$. Finally, each pixel is classified into foreground if more than $50\%$ of its average votes correspond to positives. Otherwise, the region is labeled as background. Mathematically, this is equivalent to
\begin{equation}
\hat{Y}(p_j) = \mathbbm{1}_{\mathcal{M}(p_j) > 0.5}.
\end{equation}



\section{Experiments}
To the best of our knowledge, the COCO dataset is the largest publicly available dataset for semantic/instance segmentation, with a total of $2.5$ million labeled instances in $328,000$ images \cite{Lin2016}. However, as discussed in Section \ref{sec:coco}, COCO ground truth annotations contain imprecisions intrinsic from its labeling procedure. Hence, with the purpose of minimizing the influence of dataset-specific errors, we assess the performance of FCIS and FCIS+RGR on: i) the COCO 2016 validation set; ii) the PASCAL VOC 2012 dataset; and, iii) selected video sequences of the DAVIS dataset.

For all our experiments, we defined the thresholds as follows. First, $\tau_0$ corresponds to the original detector optimal threshold. For FCIS this value is $0.4$, as reported in \cite{Li2016}. To identify high confidence foreground, we empirically selected a high confidence threshold $\tau_F$ corresponding to $1.5\times\tau_0$, hence $0.6$. As for the background detection, we configure the high confidence lower threshold as $\tau_B = 0.0$.

\begin{table*}[t]
\begin{center}
\label{tab:coco}
\begin{tabular}{lcccccccc}
\hline
 & \multicolumn{6}{c}{\textbf{COCO 2016 (val)}} & \textbf{PASCAL 2012 (val)} & \textbf{DAVIS} \\ \cline{2-9} 
 & $AP(\%)$ & $AP_{50} (\%)$ & $AP_{75} (\%)$ & $AP_S (\%)$ & $AP_M (\%)$ & $AP_L (\%)$ & $mAP (\%)$ & $IoU (\%)$ \\ \hline\hline
FCIS & 35.1 & 60.1 & 36.5 & 9.8 & 38.4 & 59.8 & 70.6 & 71.2 \\
FCIS + SPPX & 34.9 & 59.0 & 36.4 & 9.3 & 38.2 & 59.6 & 70.9 & 72.8 \\
\textbf{FCIS + RGR} & \textbf{36.9} & \textbf{60.6} & \textbf{39.3} & \textbf{11.4} & \textbf{40.7} & \textbf{60.5} & \textbf{71.1 }& \textbf{74.4 }\\ \hline
\end{tabular}
\end{center}
\caption{Comparison between results obtained by FCIS+RGR and FCIS on COCO 2016 validation set.}
\end{table*}

\begin{figure*}[b]
\begin{center}
   \includegraphics[width=\linewidth]{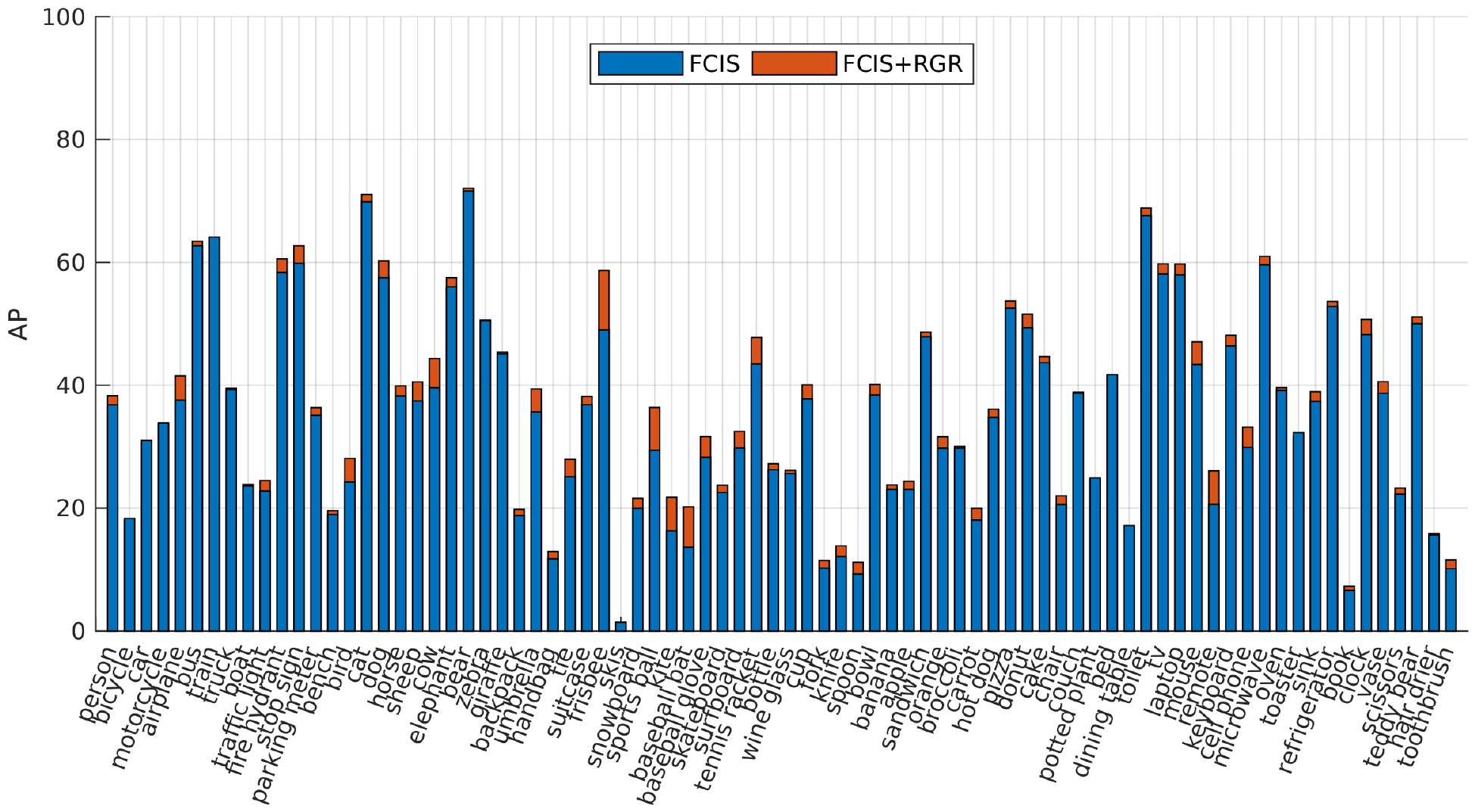}
\end{center}
   \caption{Mean average precision values obtained by FCIS and FCIS+RGR for each of the $80$ COCO categories.}
\label{fig:barcoco}
\end{figure*}

We opted for the publicly available FCIS model \cite{Li2016} trained on the MS COCO 2016 dataset, which represents the state-of-the-art approach on semantic segmentation, as the input for our refinement algorithm as well as the baseline comparison approach. We denote the combination of FCIS and the refinement module RGR as FCIS+RGR.

Since RGR works in a region growing manner that is inspired by the concept of superpixel segmentation, our performance analysis also includes FCIS+SPPX, a naive refinement method that consists of performing majority voting within superpixels generated with SNIC.

We also considered refinement using the CRF approach proposed in \cite{Krahenbuhl2012}. We evaluated the efficacy of this refinement technique on both PASCAL validation set and DAVIS selected sequences. In order to evaluate generalization capability, these experiments were performed without any dataset-specific fine-tuning of parameters. Decrease in average precision values were observed for both datasets, revealing that parameters optimization is crucial for this method to be efficient.

\subsection{COCO 2016 segmentation dataset}
\label{sec:coco}
Based on the COCO official guidelines, we evaluate the performance obtained by FCIS and FCIS+RGR on the validation set composed of $40k$ images. Table \ref{tab:coco} summarizes the performance of both methods according to the standard COCO metrics including $AP$ (averaged over $0.5:0.95$ IoU thresholds), $AP_{50}$, $AP_{75}$ $AP_S$, $AP_M$ $AP_L$ (AP at different scales). While increasing the number of true positive detections ($+0.3\%$ in $AR$), for all scenarios the naive FCIS+SPPX approach decreases the $AP$ by introducing a larger number of false positives into the final detection.

Our refinement model, RGR, on the other hand, increases the baseline FCIS overall performance by $1.8\%$. Compared to the variation of $+0.5\%$ in $AP_{50}$, the increase of $2.8\%$ in $AP_{75}$ demonstrates that RGR is particularly successful for cases where the detections obtained from the input CNN are accurately located.

In addition, Figure \ref{fig:barcoco} presents the average precision obtained for each of the $80$ COCO categories. Due to its region growing strategy based on local affinity characteristics, it is natural that RGR refinement is especially effective for objects with more uniform appearance (e.g. airplane, frisbee, baseball bat). Nevertheless, these results also demonstrate the robustness of the refinement provided by RGR, since none of the object categories observe a noticeable decrease in average precision. 

\begin{figure*}[h]
\begin{center}
   \includegraphics[width=\linewidth]{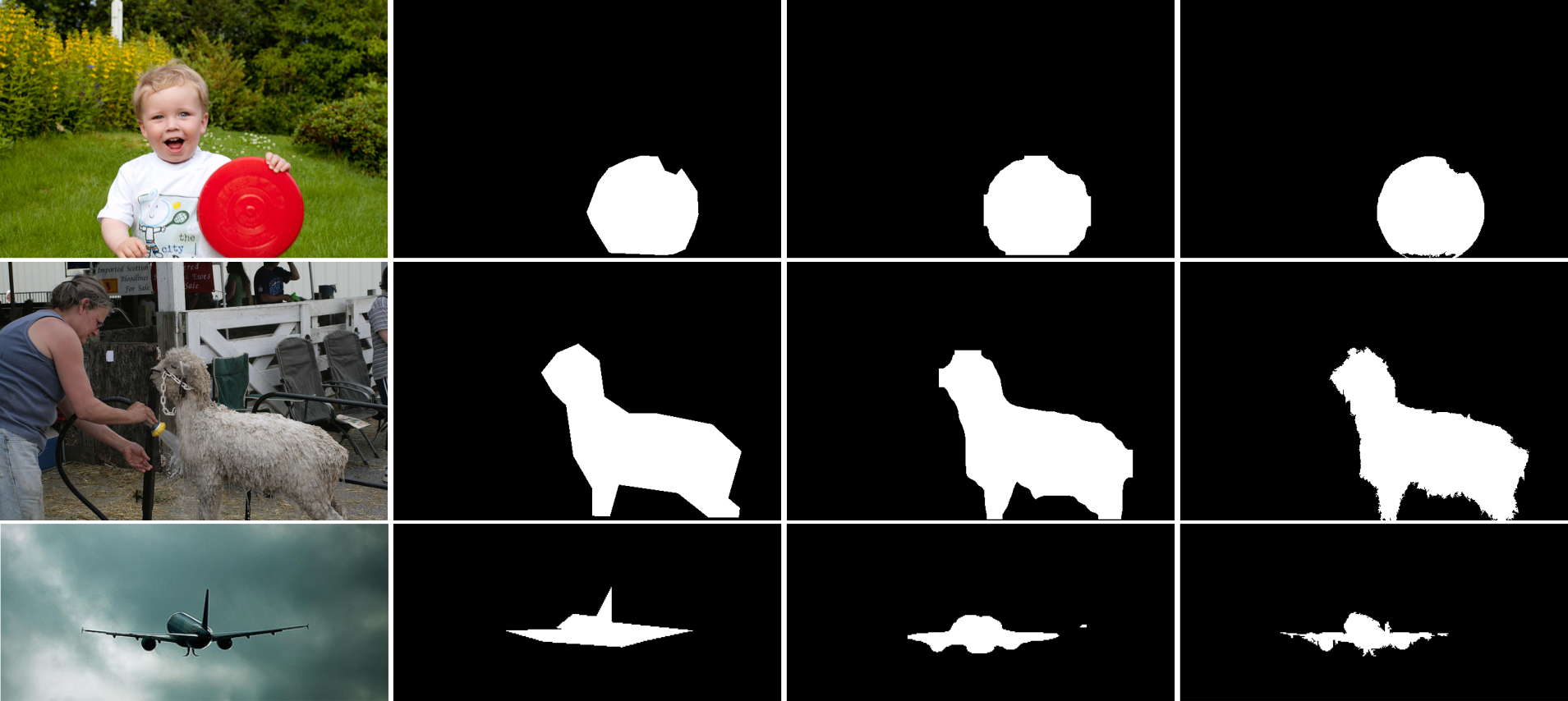}
\end{center}
   \caption{Examples of detections on the COCO dataset. From left to right: original image, ground truth, FCIS detection, FCIS+RGR. Ground truth annotations for images in the second and third rows are also examples of annotation with poor boundary adherence.}
\label{fig:cocomosaic}
\end{figure*}

A closer inspection by means of a qualitative analysis also shows that the metrics obtained might be misleading. Despite the extensive efforts and systematic approaches described in \cite{Li2016} to create such a huge dataset, for some instances the segmentation annotations provided as ground truth lack accurate adherence to real object boundaries. As consequence, a visual inspection reveals that improvements obtained through RGR refinement are not reflected in the final metrics for a significant number of images. Figure \ref{fig:cocomosaic} provides multiple examples of the imprecisions in ground truth annotations, collected from different object classes. While for the \textit{airplane} example the segmentation obtained using RGR is clearly better in qualitative terms, its overlap ($IoU$) with the ground truth is $7.7\%$ lower than the one associated with the FCIS output.


\subsection{PASCAL VOC 2012 segmentation dataset}
\begin{figure}[h]
\begin{center}
   \includegraphics[width=\linewidth]{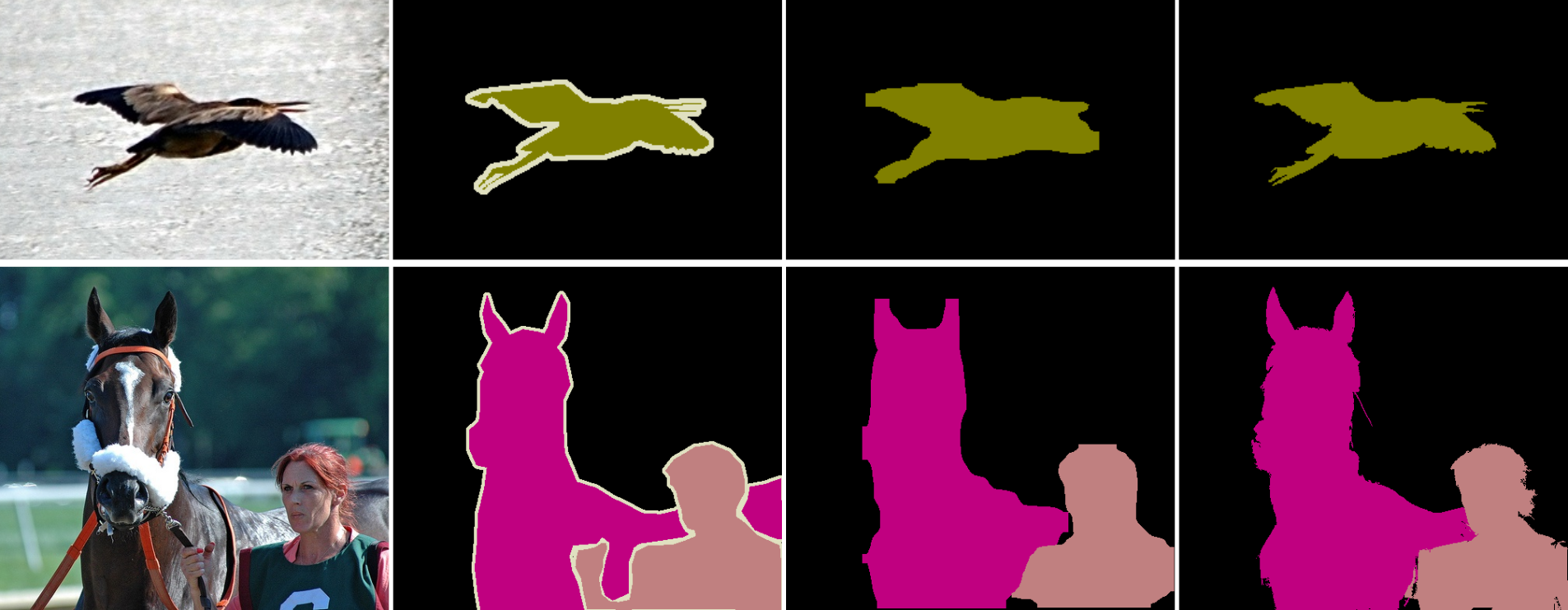}
\end{center}
   \caption{Examples of detections on the PASCAL VOC 2012 dataset. From left to right: original image, ground truth, FCIS detection, FCIS+RGR.}
\label{fig:testp}
\end{figure}

Table \ref{tab:coco} also contains a summary of the results obtained by FCIS, FCIS+SPPX and our method FCIS+RGR. The segmentation refinement generated using RGR provides final $mAP$ slightly better than both FCIS ($+0.5\%$) and the refined version using naive superpixel-wise majority voting. However, as illustrated in Figure \ref{fig:testp}, a qualitative inspection reveals that such a small difference in performance does not properly reflect the higher boundary adherence provided by the refinements produced by RGR. Although the PASCAL annotations are more pixel-accurate than the ones available for COCO, they contain $5$ pixels-wide border areas that are not consider neither object nor background, ultimately limiting the evaluation of boundary adherence.

\begin{figure}[h]
\begin{center}
   \includegraphics[width=\linewidth]{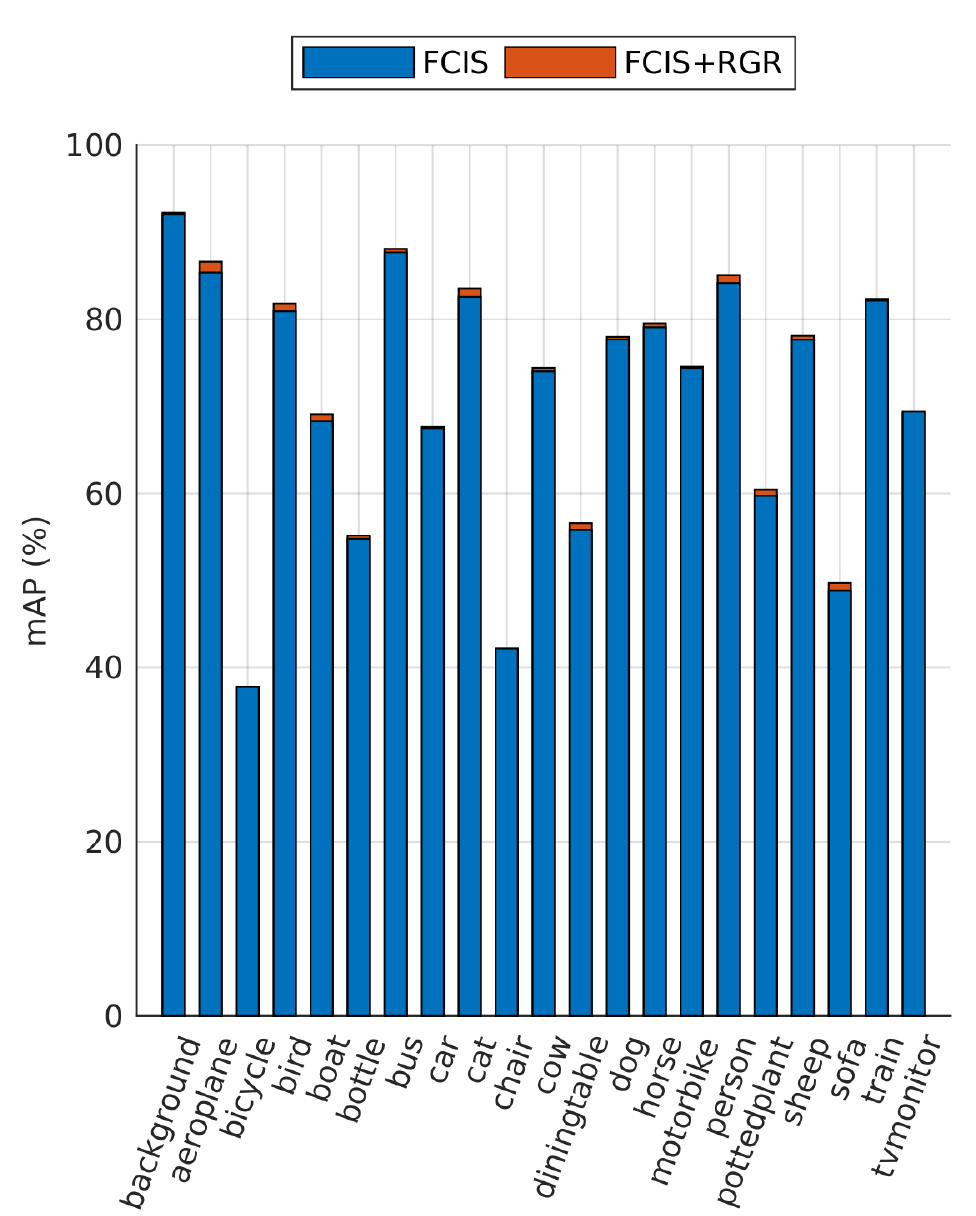}
\end{center}
   \caption{Mean average precision ($mAP$) obtained by FCIS and FCIS+RGR for each of the PASCAL VOC 2012 classes (val).}
\label{fig:bardavis}
\end{figure}

\subsection{DAVIS 2016 selected video sequences}
Given the aforementioned inaccuracies in the annotations available for the COCO and PASCAL datasets, we additionally report quantitative results on selected video sequences of the DAVIS dataset. The DAVIS dataset for Video Object Segmentation is composed by high quality video sequences \cite{Perazzi2016}, with pixel-accurate ground truth segmentation for each frame. From its $50$ original video sequences, we selected a total of $24$ sequences that contain target objects whose corresponding classes are contained within the $80$ objects categories composing the COCO dataset. Moreover, these sequences encompass $13$ different COCO classes, including classes for which FCIS+RGR did not provide significant performance improvements on the COCO evaluation (e.g. \textit{person, bicycle, boat} and \textit{bear}).

\begin{figure}[h]
\begin{center}
   \includegraphics[width=\linewidth]{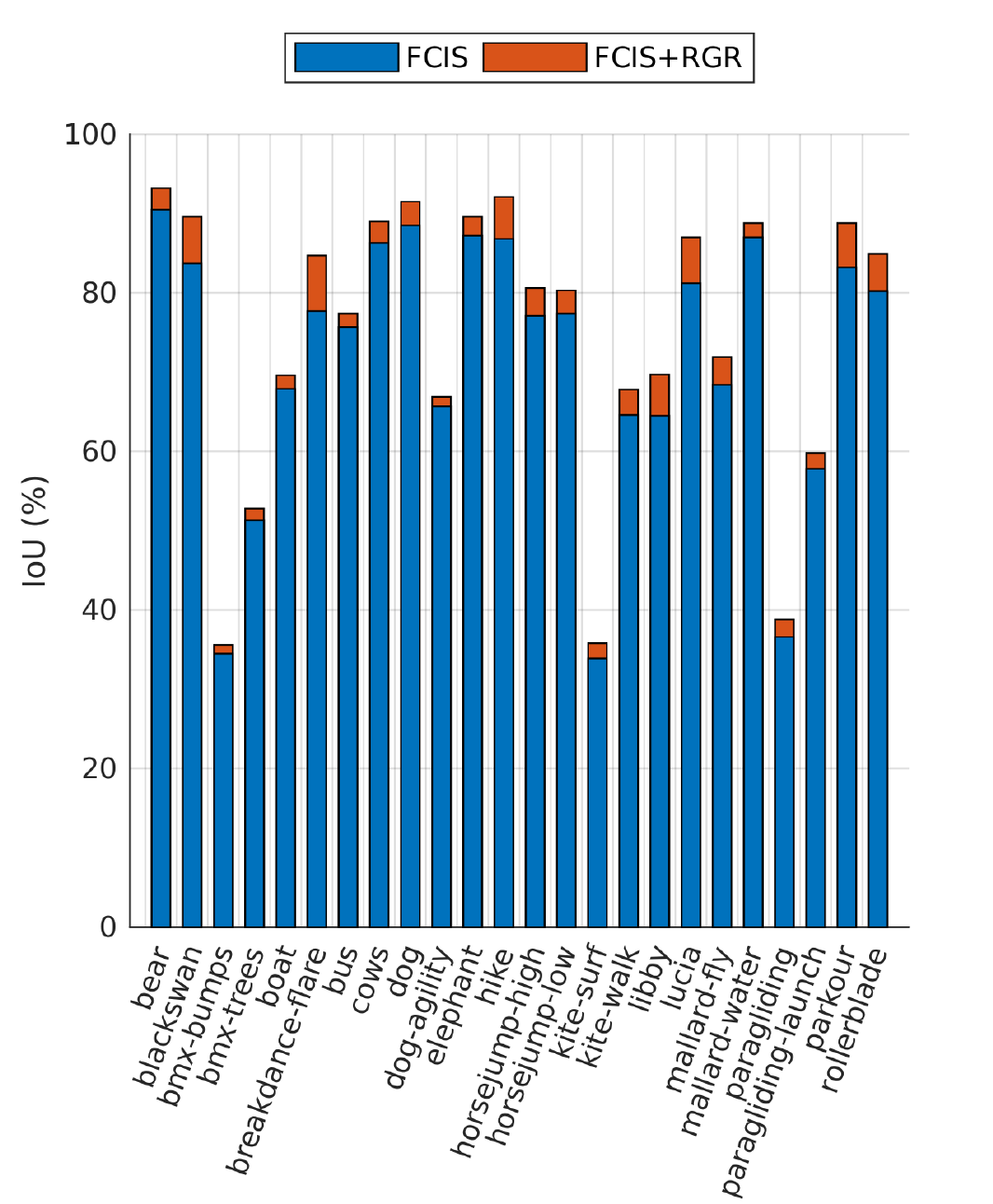}
\end{center}
   \caption{Mean IoU (Jaccard) obtained by FCIS and FCIS+RGR for each of the selected DAVIS 2016 video sequences.}
\label{fig:bardavis}
\end{figure}

As summarized in Table \ref{tab:coco}, segmentations obtained using FCIS+RGR have an average $IoU$ $3.2\%$ higher than the original FCIS ones, while $1.6\%$ better than the results obtained using a naive superpixel-wise majority voting for refinement. Official metrics for performance evaluation on the DAVIS also include a contour accuracy metric $\mathcal{F}$, which corresponds to a contour F-measure computed based on contour-based precision and recall. Segmentations refined using RGR yielded an increase of $2.8\%$ in $\mathcal{F}$, confirming its ability to improve boundary adherence.

Figure \ref{fig:bardavis} presents the results obtained for each video sequence, with FCIS+RGR providing improvements in a range from $1.1\%$ to $6.9\%$ depending on the sequence. The fact that larger quantitative improvements are obtained for the DAVIS sequences corroborates our argument that the annotations available for the COCO and PASCAL datasets provide limited information regarding boundary accuracy/adherence of segmentation methods. 

\begin{figure}[h]
\begin{center}
   \includegraphics[width=\linewidth]{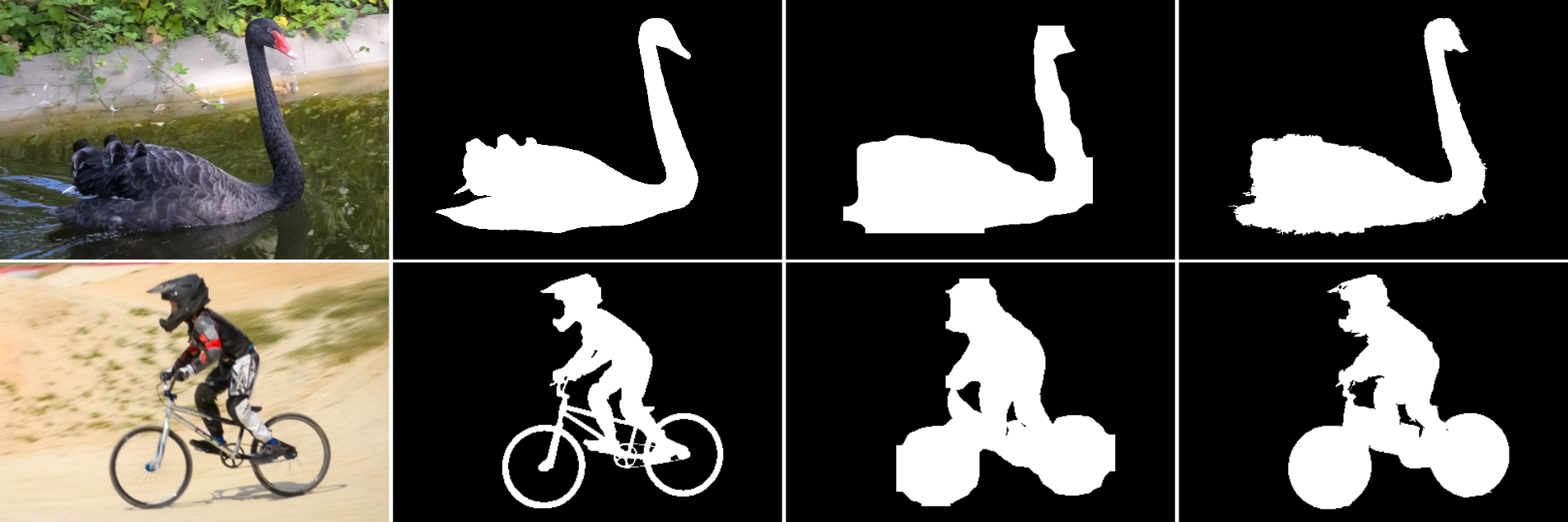}
\end{center}
   \caption{Examples of detections on the DAVIS dataset. From left to right: original image, ground truth, FCIS detection, FCIS+RGR.}
\label{fig:mosaicdavis}
\end{figure}

\subsection{Failure cases}
As demonstrated by the experimental results, the quality of the refinement obtained using RGR depends on the accuracy of the detector model used as input, especially in terms of localization. Although Monte Carlo sampling improves the robustness against high confidence false positives, errors might be propagated if the score maps collected from the detector module (CNN) contain regions with high concentration of false positives. An example of such case is found in Figure \ref{fig:mosaicdavis}. Given the high confidence of the false positives generated by the FCIS model for internal parts of the bicycle wheels, RGR is therefore unable to correct these mistakes and hence these regions remain incorrectly segmented.
\section{Conclusion}
We have presented RGR, an effective post-processing algorithm for segmentation refinement. Traditionally, the final classification step of existing methods for semantic segmentation consists of thresholding score maps obtained from DCNNs. Based on the concepts of Monte Carlo sampling and region growing, our algorithm achieves an adaptive thresholding by exploiting high confidence detections and low-level image information. Our results demonstrate the efficacy of RGR refinement, with precision increases on three different segmentation datasets. 

Moreover, we highlight limitations of existing datasets in terms of boundary adherence of available ground truth annotations. This is an intrinsic limitation of annotations procedures that rely on approximating segmentations as polygons. While the quality of the segmentation is proportional to the number of vertices selected by the user, on the other hand selecting more points increases the labeling time per object. As future work, we consider exploiting RGR for improving annotations available for existing datasets or designing an alternative annotation tool. Finally, we hypothesize that the performance of existing DCNNs for semantic segmentation could be improved by training with RGR as an additional step before computing losses. Such an arrangement could lead to a detector module that optimally interacts with RGR, identifying keypoints for the refinement process.

{\small
\bibliographystyle{ieee}
\bibliography{refs}
}

\end{document}